\documentclass{article}
\usepackage[english]{babel}

\PassOptionsToPackage{numbers, compress}{natbib} % , compress, square
\usepackage[final]{nips_2016}

\usepackage[utf8]{inputenc} % allow utf-8 input
\usepackage[T1]{fontenc}    % use 8-bit T1 fonts
\usepackage{hyperref}       % hyperlinks
\usepackage{url}            % simple URL typesetting
\usepackage{booktabs}       % professional-quality tables
\usepackage{amsfonts}       % blackboard math symbols
\usepackage{nicefrac}       % compact symbols for 1/2, etc.
\usepackage{microtype}      % microtypography

\usepackage{graphicx}
\usepackage{algorithm}
\usepackage[]{algpseudocode}
\usepackage{multirow}

\makeatletter
\renewcommand{\ALG@beginalgorithmic}{\scriptsize}
\makeatother
\makeatletter
\def\BState{\State\hskip-\ALG@thistlm}
\makeatother

\title{A Fully-Automated Pipeline for Detection and Segmentation of Liver Lesions and Pathological Lymph Nodes}

\author{
Assaf Hoogi\footnotemark[1]  \hspace{0.05cm}\footnotemark[2]
  \hspace{0.1cm}, John W. Lambert  \thanks{Both authors contributed equally to this work.}\hspace{0.15cm}\footnotemark[2]
  \hspace{0.1cm}, Yefeng Zheng\footnotemark[5] \hspace{0.1cm}, Dorin Comaniciu\footnotemark[5] \hspace{0.1cm}, Daniel L. Rubin\footnotemark[2]\\
    \footnotemark[2]\hspace{0.1cm} Departments of Radiology, Biomedical Data Science, and\\ Medicine (Biomedical Informatics Research), Stanford, CA, USA\\
  \footnotemark[5] \hspace{0.1cm}Medical Imaging Technologies, Siemens Medical Solutions USA Inc., Princeton, NJ, USA\\
}

\begin{document}

\maketitle
\vspace{-0.6cm}
\begin{abstract}
  We propose a fully-automated method for accurate and robust detection and segmentation of potentially cancerous lesions found in the liver and in lymph nodes. The process is performed in three steps, including organ detection, lesion detection and lesion segmentation. Our method applies machine learning techniques such as marginal space learning and convolutional neural networks, as well as active contour models. The method proves to be robust in its handling of extremely high lesion diversity. We tested our method on volumetric computed tomography (CT) images, including 42 volumes containing liver lesions and 86 volumes containing 595 pathological lymph nodes. Preliminary results under 10-fold cross validation show that for both the liver lesions and the lymph nodes, a total detection sensitivity of 0.53 and average Dice score of $0.71 \pm 0.15$ for segmentation were obtained.
\end{abstract}
\vspace{-0.5cm}
\section{Introduction}
\vspace{-0.3cm}Currently, oncologists rely upon the manual measurement of lesions to assess treatment response using criteria such as Response Evaluation Criteria in Solid Tumors (RECIST) \cite{10.1148/rg.282075068.Review}. Due to the laborious nature of RECIST assessment, it is only applied to~$5\%$ of all cancer patients who are enrolled in clinical trials. Automated RECIST can be used for better assessment of treatment response and aggregate evidence to new, alternative biomarkers to RECIST \cite{10.1038/nrclinonc.2009.63.Review}.

\section{Related Work}
\vspace{-0.3cm}Previous work directed toward automatic lymph node detection is limited \cite{10.1117/12.811101} \cite{Kitasaka:2007:AEL:1775835.1775884}. A special filter was used in \cite{Kitasaka:2007:AEL:1775835.1775884} to detect lymph nodes. This minimum directional difference (Min-DD) filter is constructed with the assumption that lymph nodes have uniform intensity, which is not always the case. The Min-DD filter method \cite{Kitasaka:2007:AEL:1775835.1775884} was improved in \cite{10.1117/12.811101} by adding a Hessian-based blobness measure for reducing false positives. Several automatic algorithms have been proposed for liver lesion detection and segmentation, including combinations of adaptive multi-thresholding and morphological operators 
\cite{:/content/aapm/journal/medphys/31/9/10.1118/1.1782674} or k-means clustering on mean shift filtered
images \cite{Massoptier2008}. However, these histogram-based methods require a good contrast between lesions and parenchyma. Other techniques, such as AdaBoost, have been used in both semi-automatic approaches \cite{1698917} and in automatic settings to classify image textures
\cite{Pescia2008AutomaticDO}. In another approach, Shimizu \textit{et al.}\cite{SHIMIZU+NARIHIRA+FURUKAWA+KOBATAKE+NAWANO+SHINOZAKI2008} trained two AdaBoost classifiers with a set of statistical and gradient features, as well as with features based on a convergence index filter that enhances blob-like structures. Our work is innovative due to the following:

1) To our knowledge, we are the first to introduce a fully automated end-to-end pipeline that is generalizable across multiple organs.

2) The method is accurate and robust in its analysis of highly diverse lesions by effectively handling lesions with low contrast and substantial heterogeneity found within various organs.
\section{Methods}
\vspace{-0.3cm}
\subsection{Organ Detection}
\begin{figure}[h]
\begin{center}
\includegraphics[width=3in]{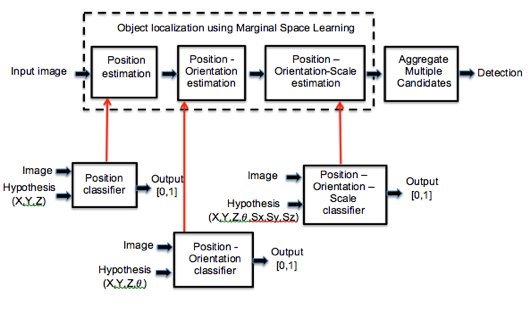}
\caption{The pipeline for organ detection }
\label{fig:PipelineSiemensOrganDetection}
\end{center}
\end{figure}

\vspace{-0.3cm}The organ detection step is based on Marginal Space Learning \cite{10.1109/CVPR.2008.4587393} (See Fig.~\ref{fig:PipelineSiemensOrganDetection}). In Marginal Space Learning, the detection is performed by using a sequence of learned classifiers, starting with classifiers with a few parameters (e.g., organ position without orientation and scale) and ending with a classifier that models all the desired organ parameters (e.g., position, orientation and scale). Each learned classifier is a Probabilistic Boosting Tree (PBT) with 3D Haar and steerable features and is trained via AdaBoost. The output probability can be within the range $[0\hspace{0.1cm} 1]$. The overall system architecture consists of two layers: 1) a Discriminative Anatomical Network (DAN), and 2) a database-guided segmentation module. The DAN supplies an estimate regarding the scale of the patient and the portion of the visible volume. Furthermore, it detects a set of landmarks for navigating the volume. The database-guided segmentation module uses the output of the DAN for the detection of the position, orientation, and scale of the organs visible in the given portion of the  volume. 
\subsection{Lesion Detection }
\vspace{-0.3cm}Liver lesions and lymph nodes are detected based on cascaded classifications \cite{DBLP:journals/tmi/BarbuSXLZC12}. The detection pipeline has three steps. First, we exploit 3D Haar-like features using an integral image of the organ sub-volume of interest. Next, we use the AdaBoost classifier to perform feature selection and classifier training simultaneously. This is especially preferred since our Haar-like feature pool contains many weak features. Third, we use self-aligned image features to train another classifier to prune the candidates. The features are based on rays cast in 14 directions from each 3D candidate. The positions with maximum gradients along each ray are determined, and 24 local image features (based on intensity, gradient, and orientation) are extracted at each position\cite{DBLP:journals/tmi/BarbuSXLZC12}. Similarly, AdaBoost is used to train the classifier. The lesion detection algorithm is presented in Algorithm ~\ref{lesiondetection}:
\vspace{-0.3cm}
\begin{algorithm}
\caption{}\label{lesiondetection}
\begin{algorithmic}[1]
\Procedure{Lesion detection}{}
\State \textbf{Input: }CT Volumes of lymph nodes and liver lesions
\State Extract sub-volumes of interest ($V$)
\For{each sub-volume $V$}
    \State Obtain initial candidates $C_0$ as all locations with intensity within [-100, 200] HU %\texttt{<do stuff>}
    \State Of the $C_0$ candidates, keep the candidates $C_1$ that pass the \textbf{3D Haar detector}
    \State Of the $C_1$ candidates, keep the candidates $C_2$ that pass the \textbf{self-aligning detector}
        \For{each $c_i = (x_i,y_i,z_i) \in C_2 $}
            \State Obtain a rough segmentation $S_i$ with center at $c_i$
            \State Obtain a score $p_i$ from the detector based on the features extracted in $S_i$
        \EndFor
        \State Discard from $C_2$ all $c_i$ with $p_i < \tau $ (defined threshold), obtaining $C_3$
        \State Call non-maximal suppression \cite{DBLP:journals/tmi/BarbuSXLZC12} on $C_3$ to obtain the detected lesions
    \EndFor
    \State \textbf{Output:} Set $D$ of detected lesions and lymph nodes
\EndProcedure
\end{algorithmic}
\end{algorithm}

\vspace{-2.0em}
\vspace{-2.0em}
\subsection{Lesion Segmentation}
\vspace{-0.3em}We conduct lesion segmentation by incorporating a convolutional neural network (CNN) and active contours. The performance of the level set segmentation depends not only upon the local region in which the energy statistics are calculated, but also upon the weighting parameters of the energy functional. The strength of the proposed method is its generalizability and ability to overcome a variety of segmentation challenges. Thus, our novel technique utilizes the benefits of both approaches and overcomes their limitations to achieve significantly better results than either method alone.
\subsubsection{Adaptive Region}
\vspace{-0.3em}In \cite{DBLP:journals/corr/HoogiBCHSNR16}, the authors proposed an iterative approach to calculate the adaptive size of the local window. The algorithm is applied for each point, at each iteration of the segmentation process, and for each lesion in the image database. The adaptive window is calculated using the lesion scale and its texture. Texture analysis is accomplished by extracting Haralick image features (e.g. contrast, homogeneity) from a second order statistics model, namely, from gray-level co-occurrence matrices (GLCM). Our method incorporates both global and local texture in a single hybrid model. \vspace{-0.3em}
\subsubsection{Adaptive Parameters of the Energy Functional}
\vspace{-0.3em}The functional parameters play a key role in the direction and magnitude of curve evolution. In \cite{Hoogi2016} the authors present a method to adaptively calculate those parameters. First, a convolutional neural network (CNN) is used to estimate the location of the Zero Level Set contour (ZLS) relative to the lesion. Three possible locations are considered: outside the lesion, near the lesion boundaries, or inside the lesion boundary. The CNN outputs a probability for each of three classes: inside the lesion and far from its boundaries ($p_1$), close to the boundaries of the lesion ($p_2$), or outside the lesion and far away from its boundaries ($p_3$). In the second step, we use the CNN probabilities of the three classes to set the weighting parameters $\lambda_1$ and $\lambda_2$. In general, $\lambda_1$ tends to contract the contour while $\lambda_2$ tends to expand it. These parameters are calculated using the following equations:
\[\small \lambda_1 =exp( \frac{1+ p_2+ p_3 }{1+p_1})\hspace{2cm}\lambda_2 =exp (\frac{1+p_1 + p_2}{1+p_3})
\]
The authors propose a generalized architecture for the CNN, which is composed of two convolutional layers followed by two fully connected layers, including the final three-class output layer. Both convolutional layers use 5 x 5 kernels, as this size outperformed smaller kernels. %We set the depths of the first and second convolutional layers to 20 and 100 pixels, respectively. 
Each convolutional block of our CNN is composed of three layers: a filter bank layer, a nonlinearity layer composed of Leaky Rectified Linear Units (ReLU), and a max-pooling layer.\vspace{-0.4em}
\section{Experimental Data}
\vspace{-1.0em}We obtained image data from The Cancer Genome Atlas (TCGA) Liver Hepatocellular Carcinoma data collection, comprising 42 3D CT volumes of liver lesions, and from The Cancer Imaging Archive (TCIA), comprising 86 3D CT volumes containing pathological lymph nodes in the abdomen. A slice thickness of 2.5 mm and an average pixel spacing of 0.894 mm were used. The cases are challenging due to the low contrast characteristics of the images and to the varying sizes, poses, shapes and sparsely distributed locations of cancer lesions. The detection step provides two points representing the corners of the lesion bounding box. Those two points, which are considered to be the lesion diameter, were used to generate the initial circular ZLS contour. For lymph nodes, the TCIA offers a complete radiologist annotation of all lymph nodes. Therefore, we can calculate the sensitivity and false positive rates automatically. However, no complete annotation of the liver lesions exists. Consequently, we visually analyzed all detections, manually counting the quantity of false positives/negatives. In CT slices that had been marked as containing liver lesions, we evaluated the segmentation by calculating the Dice criterion between the automated segmentation and the radiologist-annotated segmentation mask. \vspace{-0.4em}
\section{Results}
\vspace{-1.0em}For combined data sets of 595 lymph nodes and 42 liver lesions,  a detection sensitivity of 0.53 was obtained. For the segmentation phase, only the true positive cases with valid ground truth have been analyzed. Table 1 shows the results of our analyses. The automated segmentation was applied twice, as follows. First, by using a ZLS contour that was initialized by the 2 automatically detected points (average Dice between the automated and the manual contours was $0.71 \pm 0.15$). Second, by using 2-points that were obtained by a manual detection (average Dice of $0.77 \pm 0.04$).  Detection and segmentation examples for different challenging lesions are presented in Figure 2.
\begin{table}[h]
\caption{Detection and segmentation results for analysis of liver lesions and lymph nodes}
\small
   \centering
\begin{tabular}{|c|c|c|c| }
    \hline
     \textbf{Task} & \textbf{Input} & \textbf{Lymph Nodes} & \textbf{Liver} \\ \hline 
    \multirow{1}{*}
    {\textbf{Detection (Sensitivity)}} & {CT Volumes} & {0.51 } & {0.85}\\
 \hline
    \multirow{2}{*}{\textbf{Segmentation (Dice) }} & Automated Lesion Detection &  $0.7111 \pm 0.159$  &  {$0.669 \pm 0.203 $}\\
    & Manual Lesion Detection  & $0.768 \pm 0.033$ & 0.815 $\pm$ 0.139 \\
 \hline
\end{tabular}
\end{table}
\begin{figure}[h]
\centering
\includegraphics[width=2.5cm]{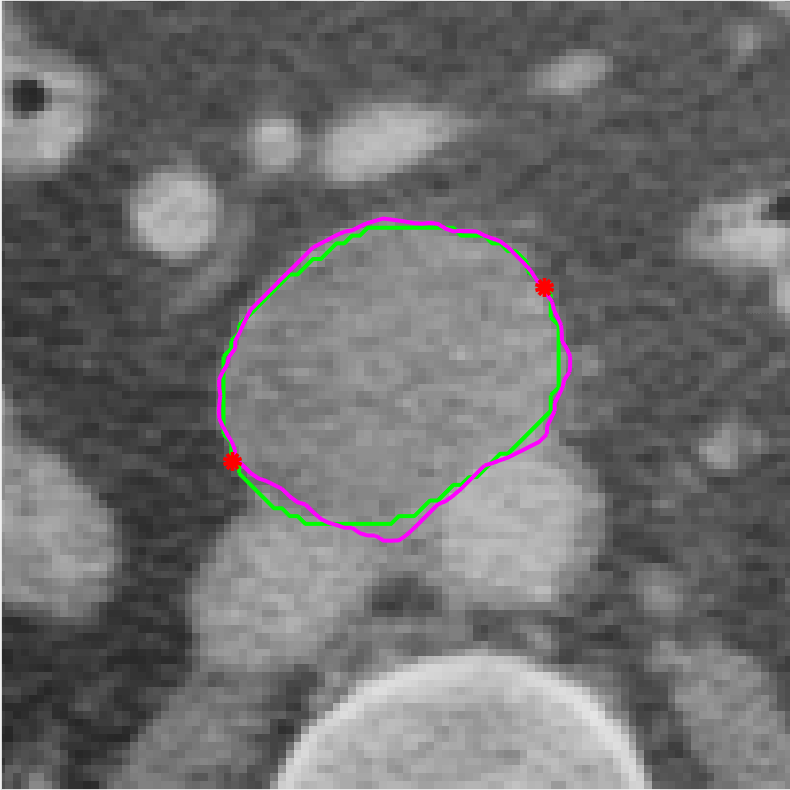}
\includegraphics[width=2.5cm]{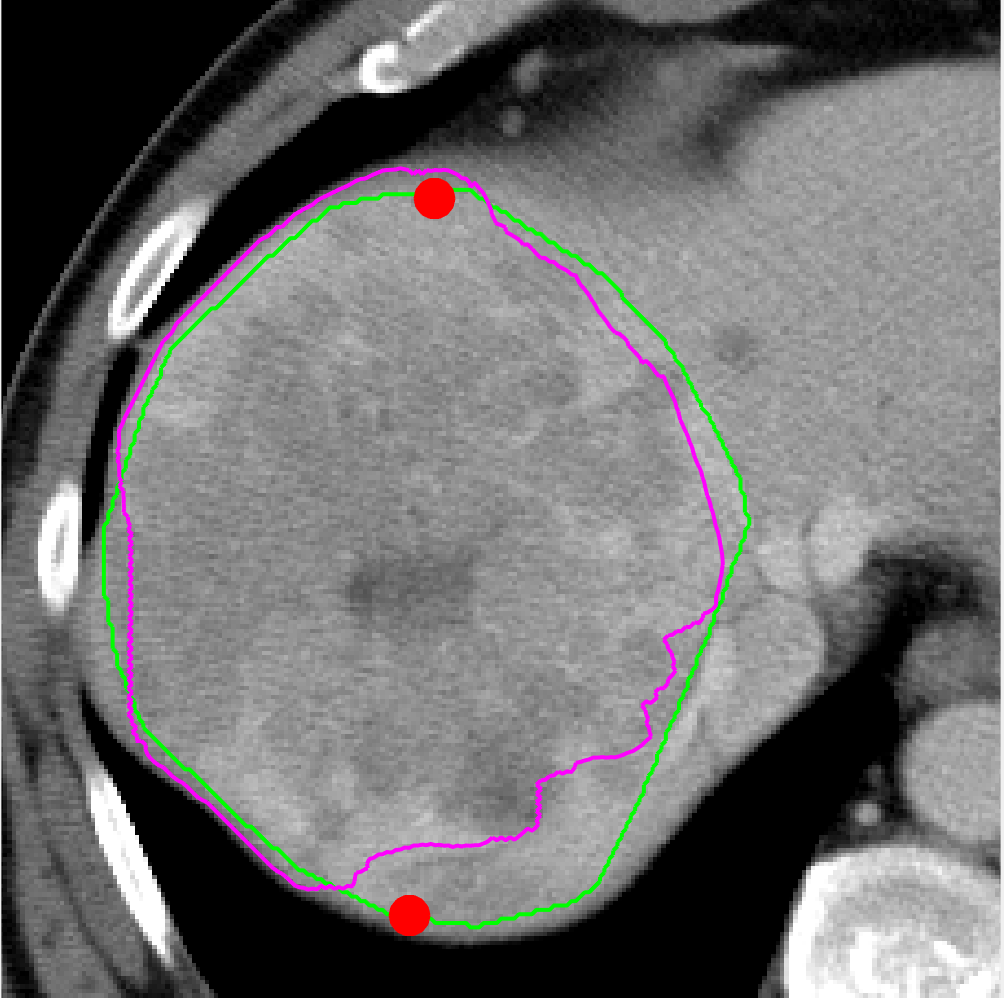}
\includegraphics[width=2.5cm]{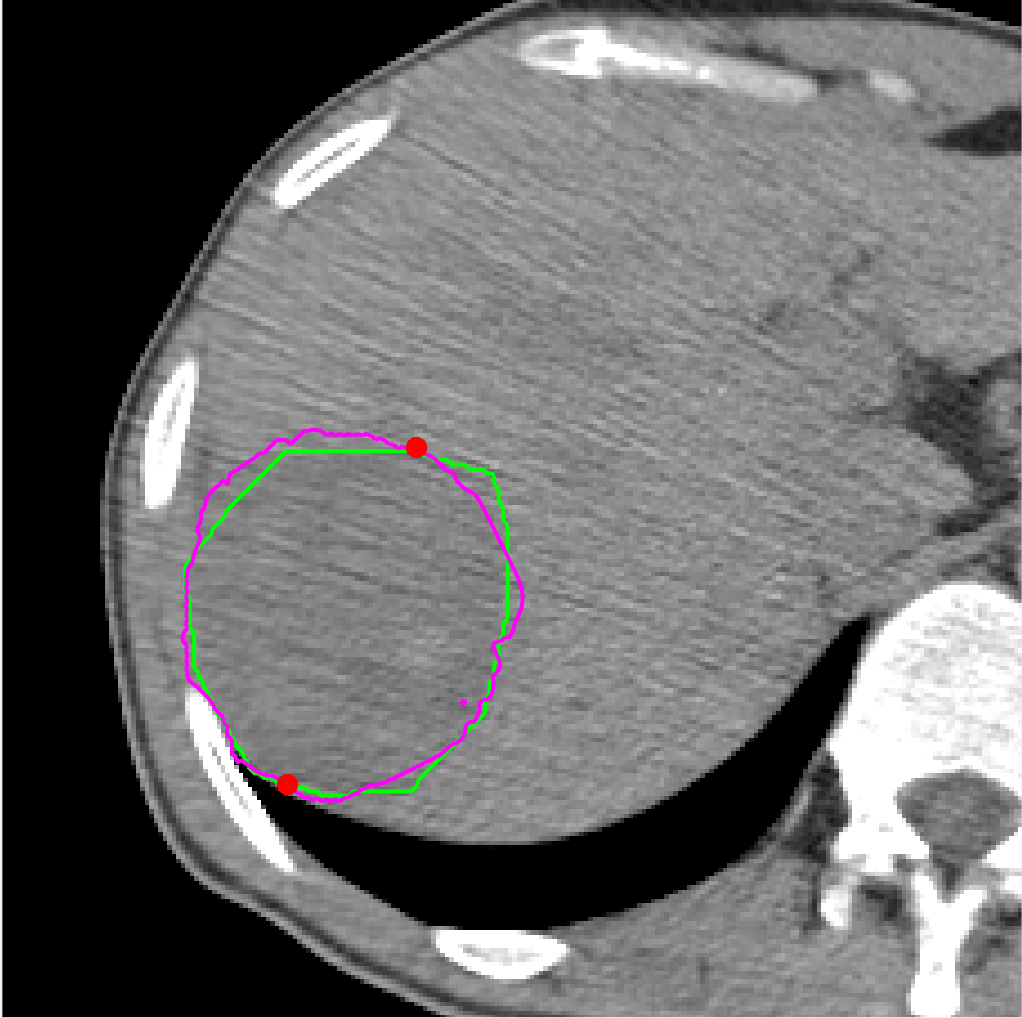}
\vspace{-0.7em}
\caption{\textbf{Left:} A pathological lymph node (Dice=0.955). \textbf{Center:} Highly heterogeneous hepatic lesion (Dice=0.913). A board-certified radiologist stated that for most parts of the lesion, the automated contour is more accurate than the manual one. \textbf{Right:} Low contrast liver lesion (Dice=0.945). Detected points - red, Manual segmentation - green, automated segmentation - magenta. }
\label{fig:figure1}
\vspace{-1.0em}
\end{figure}
\vspace{-1.0em}\section{Conclusions and Future Work}
\vspace{-1.1em}In this work, we present a novel, fully automated pipeline for the detection and segmentation of liver lesions and pathological lymph nodes. We couldn’t directly compare the literature since methods were often validated on different datasets. We believe there is room for improvement on pathological lymph node detection if we can collect enough annotated training data to leverage the recent progress in deep learning. This will be our future work. Furthermore, we note that the detection sensitivity of lymph nodes is significantly worse than that of the liver. Lymph nodes are far more challenging to detect than liver lesions for two reasons. First, liver lesions occur only inside a liver. We have developed a robust algorithm to segment the liver from a CT scan, which provides a strong constraint for liver lesion detection. However, lymph nodes appear almost everywhere inside human body; therefore we did not apply any spatial constraint during detection. Second, lymph nodes are normally small and have an intensity very close to the surrounding tissues. Liver lesions are often more easily discernible from normal liver tissue. 
\\ \vspace{-0.7em} \\We note that even when the lymph nodes boundaries are unclear around boundaries, or even when the liver lesion is very heterogeneous, our presented method can deal appropriately with both of those challenges. We also observe that the analysis of the lymph nodes is more robust than the analysis of the liver lesions, due to 1) a larger number of training cases and 2) a more accurate manual segmentation of those cases. Because the liver lesion cohort is much smaller and includes a higher percentage of challenging cases for radiologists, the training set is less accurate.   
Future work will include the extension of our current training set. We will also use a larger cohort with higher lesion diversity, including lesions in the colon, ovaries, kidneys, lymph nodes, and other organs. Moreover, additional manual markings will be used in order to obtain a more accurate evaluation and training set for our technique.\vspace{-0.8em}
\section{Acknowledgments}
\vspace{-0.9em}This work was supported in part by grants from the National Cancer Institute, National Institutes of Health, U01CA142555, 1U01CA190214, and 1U01CA187947.\vspace{-0.6em}
\begingroup
    \setlength{\bibsep}{0pt plus 0.3ex}
    {\scriptsize\bibliography{references}}
\endgroup
% \bibliography{references}

% \medskip
% \small
\end{document}